\documentclass[10pt,twocolumn,letterpaper]{article}

\usepackage{cvpr}              

\usepackage{booktabs}   
\usepackage{multirow}
\usepackage{siunitx}
\usepackage{bm}

\usepackage{algorithm}
\usepackage[noend]{algpseudocode}

\usepackage{stfloats}   

\definecolor{cvprblue}{rgb}{0.21,0.49,0.74}
\usepackage[pagebackref,breaklinks,colorlinks,allcolors=cvprblue]{hyperref}

\def\paperID{16090} 
\def\confName{CVPR}
\def\confYear{2026}

\title{SCoCCA: Multi-modal Sparse Concept Decomposition via Canonical Correlation Analysis}

\author{
Ehud Gordon\thanks{Equal contribution.} \quad Meir Yossef Levi$^{*}$ \quad Guy Gilboa \\
Viterbi Faculty of Electrical and Computer Engineering\\
Technion -- Israel Institute of Technology, Haifa, Israel\\
{\tt\small \{ehud.gordon,me.levi\}@campus.technion.ac.il; guy.gilboa@ee.technion.ac.il}
}

\begin{document}
\maketitle
\begin{abstract}
Interpreting the internal reasoning of vision-language models is essential for deploying AI in safety-critical domains. Concept-based explainability provides a human-aligned lens by representing a model's behavior through semantically meaningful components. However, existing methods are largely restricted to images and overlook the cross-modal interactions. Text–image embeddings, such as those produced by CLIP, suffer from a modality gap, where visual and textual features follow distinct distributions, limiting interpretability. Canonical Correlation Analysis (CCA) offers a principled way to align features from different distributions, but has not been leveraged for multi-modal concept-level analysis.
We show that the objectives of CCA and InfoNCE are closely related, such that optimizing CCA implicitly optimizes InfoNCE, providing a simple, training-free mechanism to enhance cross-modal alignment without affecting the pre-trained InfoNCE objective. Motivated by this observation, we couple concept-based explainability with CCA, introducing \emph{Concept CCA (CoCCA)}, a framework that aligns cross-modal embeddings while enabling interpretable concept decomposition. We further extend it and propose \emph{Sparse Concept CCA (SCoCCA)}, which enforces sparsity to produce more disentangled and discriminative concepts, facilitating improved activation, ablation, and semantic manipulation. Our approach generalizes concept-based explanations to multi-modal embeddings and achieves state-of-the-art performance in concept discovery, evidenced by reconstruction and manipulation tasks such as concept ablation. 
\end{abstract}    
\section{Introduction}
\label{sec:intro}
Developing transparent and trustworthy neural networks remains a major challenge for deploying learning systems, particularly in safety-critical domains such as autonomous driving \cite{zablocki2022explainability} and medical decision-making \cite{amann2020explainability}. Concept-based explainability (C-XAI) offers an interpretable framework for analyzing deep representations through human-understandable units, termed \emph{concepts}. Rather than relying on pixel-level saliency or feature attribution, C-XAI decomposes internal activations into disentangled, semantically coherent components that align naturally with human perception. As modern learning systems increasingly integrate multiple modalities, analyzing how multimodal learning organizes and shares conceptual structure becomes imperative. However, existing efforts have largely focused on the visual domain, leaving open the question of how concept-based explanations can be extended to multimodal networks that jointly learn from text, images, and beyond. 

\begin{figure}[t]
    \centering
    \includegraphics[width=0.99\linewidth]{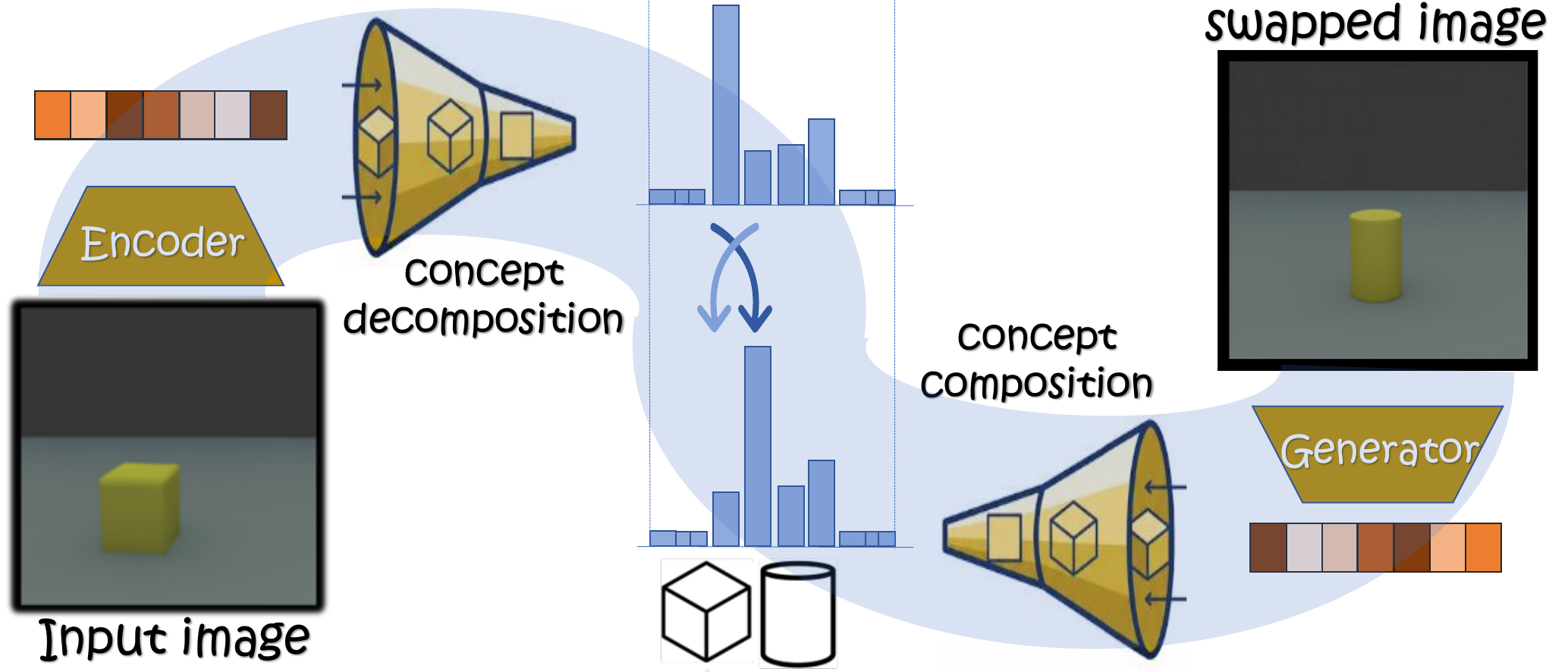}
    \caption{\textbf{Concept Swapping.} Beyond explainability, concept decomposition enables controllable manipulation. Using SCoCCA, an embedding can be decomposed into interpretable concepts (e.g., cube and cylinder), their magnitudes swapped, and the modified embedding recomposed to synthesize an image reflecting the swapped concepts.}
    \label{fig:teaser1}
\end{figure}

C-XAI has been extensively explored through approaches such as Concept Bottleneck Models \cite{pmlr-v119-koh20a} and their extensions \cite{yuksekgonul2022post, chauhan2023interactive, kim2023probabilistic}, as well as Concept Activation Vectors \cite{kim2018interpretability} and their numerous variants \cite{moayeri2023text2concept, zhang2021invertible, pfau2021robust}. These approaches, along with more recent formulations such as Varimax \cite{zhao2025quantifying}, remain confined to the image domain and fail to generalize to multi-modal settings, thereby overlooking valuable cross-modal information.
More recently, several efforts have adapted Sparse Autoencoders (SAEs) to enhance the interpretability of vision and vision–language models (VLMs) \cite{fel2025archetypal, stevens2025sparse, lim2024sparse, joseph2025steering}. Together with SpLiCE \cite{bhalla2024interpreting}, these works advanced concept decomposition in joint text–image embeddings, showing promising results. However, all existing methods either rely solely on the visual modality or overlook the inherent \emph{modality gap} \cite{NEURIPS2022_702f4db7} present in CLIP-like architectures. CLIP representations are known to exhibit a modality gap, where image and text features follow distinct distributions with mismatched geometric and probabilistic structures \cite{betser2025whitenedclip, levi2025double}, ultimately constraining both interpretability and concept reconstruction quality.

\begin{figure*}[t]
    \centering
    \includegraphics[width=0.8\linewidth]{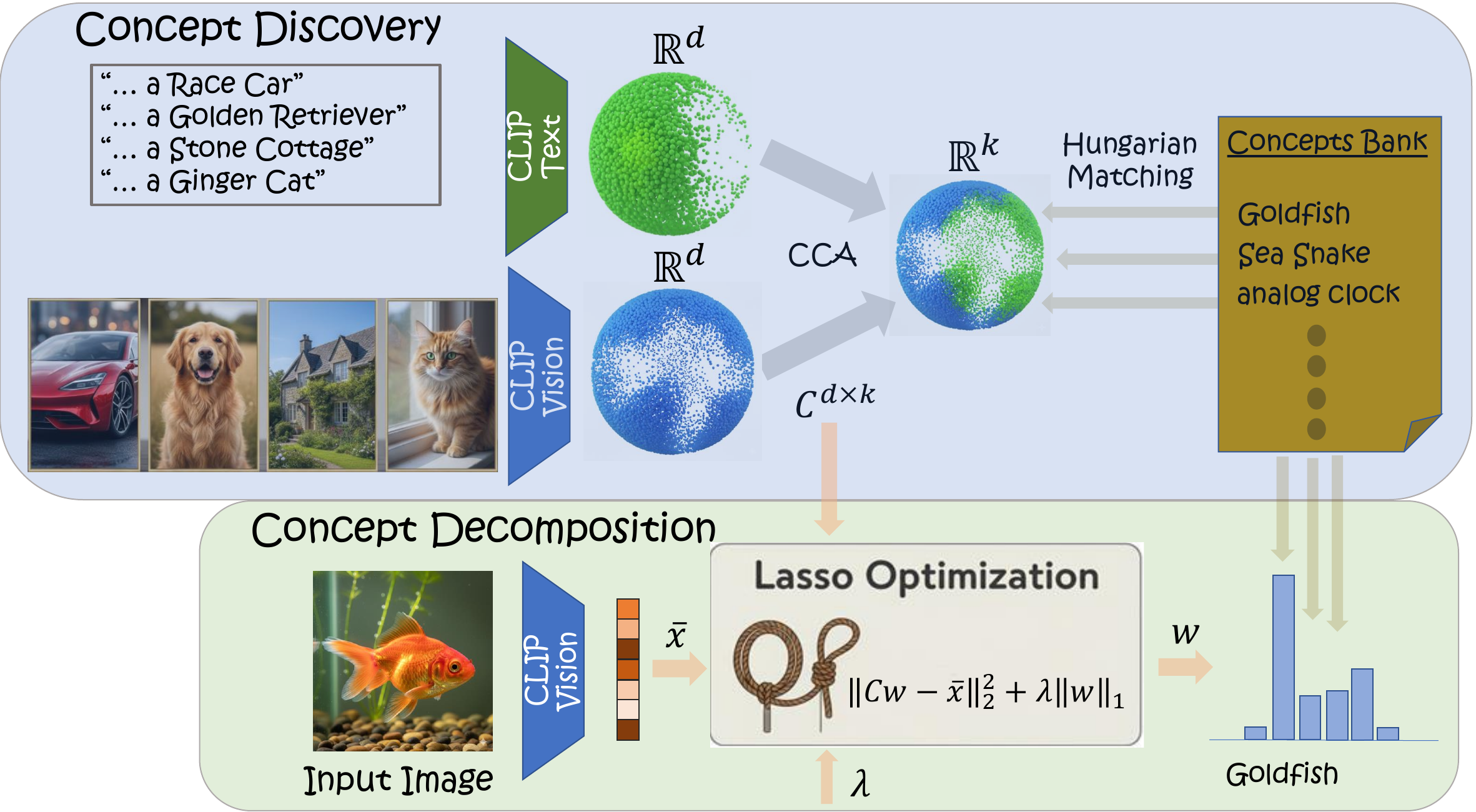}
    \caption{\textbf{Method Overview.} In the \textit{Concept Discovery} phase, text and image embeddings are aligned via Canonical Correlation Analysis (CCA) to form a shared latent space. The Hungarian algorithm establishes a one-to-one correspondence between each concept vector and its most relevant item in the concept bank. In the \textit{Concept Decomposition} phase, new embedding is decomposed into concepts by solving a Lasso optimization using the matrix $\mathbf{C}$ and the discovered associations.}
    \label{fig:teaser2}
\end{figure*}

An orthogonal line of research builds on \emph{Canonical Correlation Analysis (CCA)} \cite{hotelling1992relations}, a well-grounded mathematical framework for aligning distinct observations. These CCA-based approaches,
like multi-modal latent alignment schemes \cite{yacobi2025learning, sun2019multi, Hadgi_2025_CVPR}, emphasize correlation maximization between modalities rather than interpretability or concept analysis. While effective for cross-modal alignment, they overlook the goal of concept-level decomposition. In this work, we show that the CCA and InfoNCE \cite{oord2018representation} objectives are closely related: optimizing CCA correlates with optimizing the alignment component of the InfoNCE loss, making it a natural choice for aligning networks pre-trained with InfoNCE, such as CLIP \cite{radford2021learning}. 

Motivated by this, we propose a method termed \emph{Concept CCA (CoCCA)}, a framework that unifies the interpretability objective of concept-based explainability with the statistical alignment power of CCA. Furthermore, to enhance concept separation and interpretability, we integrate sparsity principles inspired by C-XAI into the CCA formulation, proposing \emph{Sparse Concept CCA (SCoCCA)}, yielding enhancement in concept decomposition. This sparse variant provides a more sharp and discriminative concept basis, enabling better concept activation, ablation, and swapping as demonstrated in Tab. \ref{tab:unified-comparison-bold}.
Our contributions are threefold:
\begin{itemize}
    \item We extend C-XAI to shared text–image embeddings, providing a unified framework that generalizes naturally to future multimodal foundation models.
    \item We establish a novel analytical link between CCA and the InfoNCE alignment loss, providing a training-free mechanism enables robust cross-modal concept decomposition.
    \item We introduce SCOCCA (Sparse COCCA), a novel framework that enforces an explicit sparsity constraint on the concept decomposition. This mechanism achieves superior concept disentanglement, leading to state-of-the-art efficiency in concept ablation and editing tasks
\end{itemize}
\section{Related Work}
\label{sec:related}
\paragraph{Vision-only concept decomposition.}
The \textit{Concept Activation Vector} (CAV) framework introduced by TCAV~\cite{kim2018interpretability} defines directional vectors corresponding to human concepts. Extensions such as ACE~\cite{ghorbani2019towards}, ICE~\cite{zhang2021invertible}, and CRAFT~\cite{fel2023craft} automate concept discovery by clustering or factorizing activations into coherent groups, building reusable concept banks. In parallel, \emph{Concept Bottleneck Models} (CBMs)~\cite{koh2020concept} make concepts explicit via an intermediate concept prediction layer, with variants incorporating interactive feedback or memory~\cite{chauhan2023interactive,steinmann2023learning}, probabilistic formulations~\cite{vandenhirtz2024stochastic}, unsupervised or weakly supervised discovery~\cite{shin2023closer,sawada2022concept}, and adaptation to large language models~\cite{sun2024concept}. Another line of research focuses on inducing interpretable structure through low-rank projections and rotations such as PCA, SVD, and Varimax~\cite{kaiser1958varimax,zhao2025quantifying}, which reveal compact, concentrated axes for human labeling. While these approaches improve interpretability within a single modality, they all overlook the rich mutual information integrated in the multi-modality framework.

\paragraph{Multimodal concept decomposition.}
Recent work investigates the emergence and alignment of human-interpretable concept axes in vision–language embedding spaces. Methods such as SpLiCE~\cite{bhalla2024interpreting} decompose CLIP vision embeddings into sparse additive mixtures of textual concepts, enabling compositional explanations. Complementary studies examine concept discovery directly in pre-trained vision–language models: \citet{zang2024pre} show that VLMs learn generic visual attributes via their image–text interface, \citet{li2024vision} evaluate cross-modal alignment of these concepts, and \citet{lee2023language} propose language-informed disentangled concept encoders. Parallel approaches from multiview representation learning, such as Canonical Correlation Analysis (CCA)~\cite{hotelling1936cca}, deep CCA~\cite{andrew2013deep}, and sparse CCA~\cite{witten2009penalized}, learn shared subspaces across modalities. 
A notable issue in vision–language models is the modality gap~\cite{NEURIPS2022_702f4db7}, where embeddings from different modalities are spanned in disjoint, non-isotropic distributions, with distinct properties~\cite{levi2025double, betser2025whitenedclip, schrodi2024two}. While current dedicated multimodal concept decomposition methods improve cross-modal understanding, they typically neglect this non-alignment. To our knowledge, our method is the first to explicitly align modalities to better extract mutual cross-modal information.
\section{Method}
\label{sec:method}
\subsection{Concept Decomposition Framework} \label{subsec:concept-decomp-def-ch-method}
\textbf{Notation.} We follow the dictionary-learning framework for concept-based decomposition presented by \citet{fel2023holistic}. 
An encoder $f$ maps images to activations $\mathbf{x} \in \mathbb{R}^d$. 
For a set of $n$ image inputs, we denote $ \mathbf{X} \in \mathbb{R}^{n \times d}$. Similarly, for $n$ text inputs, we denote the activations by $\mathbf{Y} \in \mathbb{R}^{n \times d}$. 
From these activations we extract a set of $k$ Concept Activation Vectors \cite{kim2018interpretability} (CAVs), for each modality.
Each CAV is denoted $\mathbf{c}_i$, and $\mathbf{C} = (\mathbf{c}_1, \ldots, \mathbf{c}_k) \in \mathbb{R}^{d \times k}$ forms the concept dictionary. 
We will focus on computing concept dictionary for the image activations.

We assume a linear relationship between $\mathbf{C}$ and $\mathbf{X}$, therefore, we look for a coefficient matrix $\mathbf{W} \in \mathbb{R}^{k \times n}$ and a concept dictionary $\mathbf{C}$ s.t. $\mathbf{X}^{\top} \approx \mathbf{C}\mathbf{W} $. 
A desirable concept decomposition should satisfy the following properties:

\begin{enumerate}
    \item \textbf{Reconstruction:} The concept dictionary $\mathbf{C}$ and weights $\mathbf{W}$ should be able to estimate well the original embeddings, i.e., we would like a low value of
    \begin{equation} \label{eq:reconstruction-metric}
        \ell(\mathbf{C}, \mathbf{W}) =  \| \mathbf{C}\mathbf{W} - \mathbf{X}^{\top} \|_F^2,
    \end{equation}
    where $\|\cdot\|_F$ denotes the Frobenius norm.
    
    \item \textbf{Sparsity:} The concept coefficients should be sparse, promoting disentangled representations \cite{mairal2014sparse}, with the objective:
    \begin{equation} \label{eq:sparsity-metric}
        \min_{\mathbf{w}} \| \mathbf{w} \|_0 ,
    \end{equation}
    for each coefficient vector $\mathbf{w}$. 
    \item \textbf{Purity:} Each concept direction $\mathbf{c}_i$ should align with human-understandable semantics.  
    This property is quantitatively assessed by applying concept-ablation and concept-swapping, and evaluating the performance of a linear probe. See Tab. \ref{tab:unified-comparison-bold}.
\end{enumerate}

\subsection{Motivation} 
\label{subsec:motivation}
Despite CLIP being explicitly trained to align positive image–text pairs, it has been observed that its latent space exhibits a \emph{modality gap}~\cite{NEURIPS2022_702f4db7}, where image and text embeddings are linearly separable~\cite{levi2025double, schrodi2024two}. 
To better capture the shared information between modalities, it is desirable to further enhance their alignment. 
This raises a natural question: \textit{why is applying CCA sensible in the context of CLIP, which is already trained using the InfoNCE loss?} 
In the following, we elaborate on the relationship between CCA and InfoNCE, demonstrating that the two objectives are closely related.

\textbf{CCA as whitened alignment.}  
Canonical correlation analysis (CCA) seeks pairs of linear projections of \(\mathbf{X}\) and \(\mathbf{Y}\) that are \emph{maximally} correlated while remaining mutually orthogonal.  
In particular, the CCA objective is to find projection matrices \(\mathbf{U}, \mathbf{V} \in \mathbb{R}^{d \times d'}\) such that
\begin{equation} \label{eq:cca-obj-matrix}
  \max_{\mathbf{U}, \mathbf{V}} \ \operatorname{tr} \big[ (\mathbf{X} \mathbf{U})^{\top}(\mathbf{Y} \mathbf{V}) \big] 
\end{equation}
subject to
\begin{equation} \label{eq:cca-orthogonal-matrices}
  (\mathbf{X} \mathbf{U})^{\top} (\mathbf{X} \mathbf{U}) = \mathbf{I}_{d'}, \quad (\mathbf{Y} \mathbf{V})^{\top} (\mathbf{Y} \mathbf{V}) = \mathbf{I}_{d'}.
\end{equation}

\paragraph{Whitening.} Let the whitened embeddings be
\begin{equation}
\label{eq:whitening}
\widetilde{\mathbf{X}} := (\mathbf{X}-\mathbf{1}\bm{\mu}_X^\top) \mathbf{W_X}, \qquad 
\widetilde{\mathbf{Y}} := (\mathbf{Y}-\mathbf{1}\bm{\mu}_Y^\top) \mathbf{W_Y},
\end{equation}
with \(\bm{\mu}_X = \tfrac{1}{n}\mathbf{X}^\top \mathbf{1}\) and \(\bm{\mu}_Y = \tfrac{1}{n}\mathbf{Y}^\top \mathbf{1}\).
The whitening matrix \cite{kessy2018optimal} of a set of vectors \(\mathbf{X}\) is the linear transformation \(\mathbf{W}_{\mathbf{X}}\) satisfying  
\begin{equation} \label{eq:whitening-matrix-x}
  \frac{1}{n}\widetilde{\mathbf{X}}^{\top} \widetilde{\mathbf{X}} = \mathbf{I}_{d},
\end{equation}
i.e., a matrix that projects \(\mathbf{X}\) to have identity covariance and zero mean.   
Note that \(\mathbf{W}_{\mathbf{X}}\) is not uniquely determined by \eqref{eq:whitening-matrix-x}; in fact, any multiplication of a whitening matrix $\mathbf{W}$ by an orthogonal matrix yields another whitening matrix. A common solution is to choose PCA-whitening. 
Following ~\citet{jendoubi2019whitening}, we set \(\mathbf{W}_{\mathbf{X}} = \mathbf{U}\), \(\mathbf{W}_{\mathbf{Y}} = \mathbf{V}\).  
Since \(\mathbf{U}\) and \(\mathbf{V}\) satisfy Eq. \eqref{eq:cca-orthogonal-matrices}, they also satisfy the whitening condition \eqref{eq:whitening-matrix-x}.  
This recasts the CCA objective as a simultaneous whitening of both \(\mathbf{X}\) and \(\mathbf{Y}\), which can be rewritten as
\begin{equation} \label{eq:cca_as_whitened}
  \max_{\mathbf{U}, \mathbf{V}} \operatorname{tr} \big[ \widetilde{\mathbf{X}}^{\top} \widetilde{\mathbf{Y}} \big].
\end{equation}
In other words, CCA can be interpreted as maximizing the alignment between two \emph{whitened} sets of observations.

\paragraph{InfoNCE on whitened inputs.}  
The InfoNCE loss (considering one of the two symmetric directions) can be decomposed into alignment and uniformity terms \cite{wang2020understanding} and written as
\begin{equation}
    \mathcal{L}_{\mathbf{X} \to \mathbf{Y}}
    = \underbrace{-\tfrac{1}{N\tau} \mathrm{tr}(\mathbf{X}^{\top}\mathbf{Y})}_{\text{alignment}}
    + \underbrace{\tfrac{1}{N\tau}\mathbf{1}^\top \log\!\big(\exp(\mathbf{X}^{\top}\mathbf{Y})\mathbf{1}\big)}_{\text{uniformity}},
\end{equation}
where \(\exp(\cdot)\) and \(\log(\cdot)\) are applied element-wise.
Then, the InfoNCE loss on whitened embeddings becomes
\begin{equation}
\mathcal{L}_{\widetilde{\mathbf{X}}\to \widetilde{\mathbf{Y}}}^{(w)}
= \underbrace{-\tfrac{1}{N\tau} \mathrm{tr}(\widetilde{\mathbf{X}}^{\top}\widetilde{\mathbf{Y}})}_{\text{alignment}}
+ \underbrace{\tfrac{1}{N\tau}\mathbf{1}^\top \log\!\big(\exp(\widetilde{\mathbf{X}}^{\top}\widetilde{\mathbf{Y}})\mathbf{1}\big)}_{\text{uniformity}}.
\end{equation}
Hence, the alignment term of \(\mathcal{L}^{(w)}\) is proportional to the CCA objective (Eq.~\eqref{eq:cca_as_whitened}). 

\paragraph{CCA enhances InfoNCE alignment.}  
The above derivation highlights a key insight: maximizing the CCA objective implicitly optimizing the alignment InfoNCE term of whitened inputs.  
In other words, CCA can \emph{implicitly} enhance the optimization of a pretrained InfoNCE-based model 
and may be viewed as a fine-tuning strategy.
Moreover, CCA offers this benefit with an analytical \emph{closed-form} solution, avoiding the overhead and potential pitfalls of additional training phases. 

\subsection{Sparse Concept CCA (SCoCCA)}
\label{subsection:scocca}
The proposed method consists of two main phases. 
The first, the \emph{concept discovery} phase, constructs a set of interpretable Concept Activation Vectors (CAVs) organized in the matrix $\mathbf{C}$, derived from paired image–text embeddings $(\mathbf{X}, \mathbf{Y})$. 
Each CAV is associated with a human-understandable interpretation, and this phase is performed once, a priori. 
The second, the \emph{concept decomposition} phase, interprets a new, unseen image embedding by leveraging the learned dictionary $\mathbf{C}$ from the fitting phase to decompose its activation into a sparse combination of interpretable concepts, solved via the Lasso optimization procedure~\citep{tibshirani1996lasso}. 
Beyond interpretability, the process is inherently invertible, enabling, for example, selective modification of concept activations, re-composition, and image synthesis through unCLIP~\cite{ramesh2022hierarchical}, as illustrated in Fig.~\ref{fig:teaser1}. 
In the following, we elaborate on each of these two phases, which are jointly summarized in Alg.~\ref{alg:scocca-lasso}.

\subsection{Concept Discovery}
\subsubsection{Concept CCA (CoCCA)}
To uncover shared semantic structure across modalities, we extend canonical correlation analysis (CCA) into a concept decomposition framework. 
CoCCA learns projection matrices specifically projecting to dimension $k$, $\mathbf{U}, \mathbf{V} \in \mathbb{R}^{d \times k}$ that maximize the correlation between the projected embeddings, as defined by the CCA objective in Eq.~\eqref{eq:cca-obj-matrix}, subject to the orthogonality constraints in Eq.~\eqref{eq:cca-orthogonal-matrices}. 
The resulting projections $\mathbf{U}$ and $\mathbf{V}$ capture directions that are maximally aligned between the image and text embedding spaces.
Importantly, this optimization admits a closed-form analytical solution:
\begin{equation} \label{eq:cca-obj-recast-svd}
 \mathbf{U} = \bm{\Sigma}_{\mathbf{X}}^{-1/2} \mathbf{Q}_x ,  \qquad
 \mathbf{V} = \bm{\Sigma}_{\mathbf{Y}}^{-1/2} \mathbf{Q}_y,
\end{equation}
where $\mathbf{Q}_x, \mathbf{Q}_y$ are obtained via the singular value decomposition (SVD) of 
$\mathbf{M} := \bm{\Sigma}_{\mathbf{X}}^{-1/2} \bm{\Sigma}_{\mathbf{X}\mathbf{Y}} \bm{\Sigma}_{\mathbf{Y}}^{-1/2}$,
yielding $\mathbf{M} = \mathbf{Q}_x \mathbf{S} \mathbf{Q}_y^\top$.
Here, $\bm{\Sigma}_{\mathbf{X}}$ and $\bm{\Sigma}_{\mathbf{Y}}$ denote the covariance matrices of the centered CLIP image and text embeddings, respectively.
A detailed derivation of this formulation is provided in the Supp.
Finally, the concept bank is constructed from the image projections as:
\begin{equation}
\label{eq:define_c}
\mathbf{C} := \bm{\Sigma}_{\mathbf{X}} \mathbf{U}.
\end{equation}

\subsubsection{Concept Matching via the Hungarian Method} \label{subsubsec:concept-matching-hungarian}
Obtaining $\mathbf{C}$ provides a decomposition of the embedding space into concept directions; however, these directions are not yet semantically grounded. In the following, we describe how each direction in $\mathbf{C}$ is associated with a meaningful concept label (e.g., “dog,” “cat,” etc.).
Let $\mathbf{C} = [\mathbf{c}_1, \dots, \mathbf{c}_k] \in \mathbb{R}^{d \times k}$ be the learned
concept vectors, and let $\{\mathbf{\bar{x}}_i\}_{i=1}^{N} \subset \mathbb{R}^d$ be the centered CLIP image
embeddings for a labeled dataset with $M$ classes and labels $y_i \in \{1,\dots,M\}$.
For each class $j \in \{1,\dots,M\}$, we define the index set
\[
\mathcal{I}_j = \{ i \in \{1,\dots,N\} : y_i = j \},
\]
and compute the mean image embedding of that class:
\begin{equation}
    \mathbf{p}_j = \frac{1}{|\mathcal{I}_j|} \sum_{i \in \mathcal{I}_j} \mathbf{\bar{x}}_i,
    \quad j=1,\dots,M.
\end{equation}
We then stack these class prototypes into a matrix
$\mathbf{P} = [\mathbf{p}_1, \dots, \mathbf{p}_M] \in \mathbb{R}^{d \times M}$.
Next, we compute the cosine similarity matrix between concepts and class means:
\begin{equation}
\mathbf{S} = \mathbf{C}^\top \mathbf{P}, \qquad
S_{ij} = \frac{\langle \mathbf{c}_i, \mathbf{p}_j \rangle}{\|\mathbf{c}_i\|_2 \, \|\mathbf{p}_j\|_2}.
\end{equation}
The optimal one-to-one assignment between concepts and classes is obtained by maximizing the total similarity
\begin{equation}
\label{eq:solve_b}
\max_{\mathbf{B}} \sum_{i=1}^k \sum_{j=1}^M S_{ij} B_{ij},
\end{equation}
where $\mathbf{B} \in \{0,1\}^{k \times M}$ is a binary assignment matrix whose rows and columns each contain exactly one nonzero entry, ensuring a unique match between concepts and classes. This assignment is computed efficiently using the Hungarian algorithm~\cite{Kuhn1955}.

\subsubsection{Adding sparsity to CoCCA}
With the concept dictionary in hand, we proceed to analyze new, unseen image example.
In order to have interpretable, disentangled representation in concept-space, we encourage its weights $\mathbf{w}$ to be sparse. 
Given new centered image embedding $\bar{\mathbf{x}}\in\mathbb{R}^{d}$ and a dictionary $\mathbf{C} \in\mathbb{R}^{d \times k}$, we estimate coefficients $\mathbf{w} \in\mathbb{R}^{k}$ by Lasso \citep{tibshirani1996lasso}:
\begin{equation} 
\label{eq:lasso}
\min_{\mathbf{w} \in\mathbb{R}^{k}} F(\mathbf{w}) := \underbrace{\tfrac{1}{2}\, \lVert \mathbf{C} \mathbf{w}-\mathbf{\bar{x}} \rVert_{2}^{2}}_{\text{CoCCA}} + \underbrace{\lambda \lVert \mathbf{w} \rVert_{1}}_{\text{Sparsity}},
\end{equation}
with $\lambda > 0$, balancing the amount of desired sparsity with reconstruction error. We use the Iterative Shrinkage-Thresholding Algorithm (ISTA) \citep{beck2009fista}, a proximal-gradient method for composite convex problems. ISTA alternates two explicit steps with step size $\gamma$ at step $k$:
\begin{equation}
\label{eq:lasso_iterative_pro}
\begin{split}
\text{\small (gradient step)}\quad & \mathbf{y}^{(k)} \leftarrow \mathbf{w}^{(k)} - \gamma\,\mathbf{C}^{\top}\!\big(\mathbf{C}\mathbf{w}^{(k)}-\mathbf{\bar{x}}\big),\\
\text{\small (proximal step)}\quad & \mathbf{w}^{(k+1)} \leftarrow S_{\tau\lambda}\!\big(\mathbf{y}^{(k)}\big),
\end{split}
\end{equation}
where $S_{\tau}$ is the soft-threshold operator applied component-wise:
\begin{equation}
\label{eq:soft_thresholding}
\big(S_{\tau}(\mathbf{y})\big)_{i} = \operatorname{sign}(\mathbf{y}_{i})\,\max\{|\mathbf{y}_{i}|-\tau,\,0\}.
\end{equation}
We set $\gamma=\frac{1} {\lVert \mathbf{C} \rVert_{2}^{2}}$, see \citet{parikh2014proximal} for more details. The iterative process converges to the optimized sparse coefficient $\mathbf{w}$.

\begin{algorithm}[htb]
  \caption{\textbf{Sparse Concept CCA (SCoCCA)}}
  \label{alg:scocca-lasso}
  \begin{algorithmic}[1]
    \Require Paired embeddings $(\mathbf{X}, \mathbf{Y})$, number of concepts $k$, sparsity coefficient $\lambda$, unseen embedding $\mathbf{x}_0$
    \Statex
    \Statex \textbf{Concept Discovery}
    \State Center embeddings: $\bar{\mathbf{X}} \gets \mathbf{X} - \mathbf{1} \, \bm{\mu}_{X}^\top$, \quad $\bar{\mathbf{Y}} \gets \mathbf{Y} - \mathbf{1}\, \bm{\mu}_{Y}^\top$
    \State Compute projection: $\mathbf{U} \gets \bm{\Sigma}_{\mathbf{X}}^{-1/2}\mathbf{Q}_x$ \hfill (Eq.~\eqref{eq:cca-obj-recast-svd})
    \State Derive concept dictionary: $\mathbf{C} \gets \bm{\Sigma}_{\mathbf{X}}\mathbf{U}$ \hfill (Eq.~\eqref{eq:define_c})
    \State Match concepts to textual attributes: solve $\mathbf{B} \in \{0,1\}^{k \times M}$ \hfill (Eq.~\eqref{eq:solve_b})
    \State \textbf{Output:} Concept Bank $C$
    \Statex
    \Statex \textbf{Concept Decomposition}
    \State Center unseen embedding: $\bar{\mathbf{x}}_0 \gets \mathbf{x}_0 - \bm{\mu}_{X}$
    \State Optimize sparse concept activations through Lasso: 
    $\mathbf{w} \gets \displaystyle\min_{\mathbf{w}\in \mathbb{R}^k} F(\mathbf{w})$
    \hfill (Eq.~\eqref{eq:lasso})
    \State \textbf{Output:} Sparse coefficients $\mathbf{w}$ and interpretable reconstruction $\hat{\mathbf{x}} = \mathbf{C}\mathbf{w} + \bm{\mu}_X$
  \end{algorithmic}
\end{algorithm}
\begin{table*}[t]
\centering
\caption{\textbf{Comprehensive Performance Comparison.} Results comparing concept decomposition methods on subset of 500 random classes from ImageNet~\cite{deng2009imagenet}, grouped into dual-modality and single-modality approaches, and evaluated across a wide range of metrics. Best results are shown in \textbf{bold}, and second-best are \underline{underlined}. SCoCCA achieves state-of-the-art or on-par performance in Reconstruction, as well as in Purity \& Editing. Rightmost column shows CLIP zero-shot performance. Note the SCoCCA obtains CLIP-level performance in accuracy.}
\footnotesize
\begin{tabular}{l l | cc | c c c c | c}
\toprule
\multirow{2}{*}{Category} & \multirow{2}{*}{Metric} 
& \multicolumn{2}{c|}{Dual-Modality} 
& \multicolumn{4}{c|}{Single-Modality} 
& \multicolumn{1}{c}{Baseline} \\
\cmidrule(lr){3-4} \cmidrule(lr){5-8} \cmidrule(lr){9-9}
 &  & SCoCCA \textbf{(Ours)} & SpLiCE \cite{bhalla2024interpreting} 
 & TCAV\cite{kim2018interpretability} & Varimax\cite{zhao2025quantifying} 
 & NMF & K-Means & CLIP\cite{radford2021learning} \\
 
\midrule
\multirow{5}{*}{\textbf{Purity and Editing}} 
& Ablation prob. drop (\(\uparrow\)) & \underline{0.87} & 0.20 & \textbf{0.93} & 0.81 & 0.01 & 0.47 & - \\
& Target prob gain (\(\uparrow\)) & \textbf{0.95} & 0.36 & 0.10 & \underline{0.75} & 0.02 & -0.09 & - \\
& Img residual cosine (\(\uparrow\)) & \textbf{0.76} & \underline{0.70} & 0.67 & 0.67 & 0.63 & 0.74 & - \\
& Zero-shot accuracy (\(\uparrow\)) & \textbf{0.74} & 0.48 & 0.62 & \underline{0.65} & 0.12 & 0.53 & \textsl{0.75} \\
 & Zero-shot precision@5 (\(\uparrow\)) & \textbf{0.85} & 0.61 & 0.59 & \underline{0.75} & 0.29 & 0.70 & \textsl{0.85} \\
\midrule
\multirow{3}{*}{\textbf{Sparsity}} 
 & Concepts orthogonality (\(\uparrow\)) & \underline{0.93} & 0.88 & 0.85 & \textbf{1.0} & 0.23 & 0.86 & - \\
 & Energy coverage@10 (\(\uparrow\)) & 0.30 & \underline{0.89} & 0.21 & 0.33 & 0.08 & \textbf{1.0} & - \\
 & Hoyer sparsity (\(\uparrow\)) & 0.38 & \underline{0.88} & 0.28 & 0.38 & 0.13 & \textbf{1.0} & - \\
\midrule
\multirow{2}{*}{\textbf{Reconstruction}} 
 & Cosine rec. similarity (\(\uparrow\)) & \textbf{0.99} & 0.58 & 0.50 & \underline{0.90} & 0.74 & 0.84 & - \\
 & Relative $L_2$ rec. error (\(\downarrow\)) & \textbf{0.04} & 0.35 & 0.28 & \underline{0.15} & 0.43 & 0.29 & - \\
\midrule
\end{tabular}
\label{tab:unified-comparison-bold}
\end{table*}

\section{Experiments}
\label{sec:experiments}

We comprehensively evaluate the performance of SCoCCA across multiple experiments and under several metrics, assessing its \emph{purity} and \emph{editing} capabilities, \emph{sparsity} of the concept decomposition, and \emph{reconstruction} accuracy.

\noindent\textbf{Implementation Details.} All experiments are conducted using CLIP ViT-L/14 \cite{radford2021learning} as the backbone, while results on CLIP ViT-B/32 are provided in the Supp.  
Unless stated otherwise, we use a subset of 500 randomly-chosen classes from ImageNet~\cite{deng2009imagenet} as the primary dataset, where the concept bank consists of items corresponding to ImageNet classes (e.g., ``An image of a beagle''). 
To perform concept matching using the Hungarian algorithm \cite{Kuhn1955} we use the \texttt{SciPy} library.
For computing coefficient vectors $w$, as in Eq. \ref{eq:lasso}, we use the \texttt{scikit-learn}~\cite{pedregosa2011scikit} FISTA solver. 
We compare our method against single-modality methods: TCAV~\cite{kim2018interpretability}, Varimax~\cite{zhao2025quantifying}, NMF, K-Means, and CLIP~\cite{radford2021learning}; and SpLiCE~\cite{bhalla2024interpreting}, a recent dual-modality approach. 
We use the official SpLiCE implementation provided at \cite{SpLiCE2025}, and have implemented the other methods and baselines, following the original papers implementation details. 
We used the \texttt{scikit-learn} \cite{pedregosa2011scikit} implementation for the K-Means \cite{lloyd1982least} clustering method using $k$ as number of classes on the centered ImageNet\cite{deng2009imagenet} embeddings. 
We have used the multiplicative updates solver in \texttt{scikit-learn} \cite{pedregosa2011scikit} to implement the NMF \cite{lee2000algorithms} method.
All the results are summarized in Tab. \ref{tab:unified-comparison-bold}.

\subsection{Metrics}
We train a logistic regression classifier $h$ on the ImageNet training set to predict among $M$ classes. In all metrics we average on the entire dataset unless explicitly written otherwise.
\subsubsection{Concept Purity}
A key question is how well the computed concept vectors align with semantic concepts. We evaluate this through two scenarios: (1) \emph{concept ablation}, where the coefficient corresponding to a specific concept is set to zero, and (2) \emph{concept insertion}, where a concept activation value is moved into another concept entry.
These modifications yield a new coefficient vector, denoted as $\mathbf{w}^*_0$, and its corresponding reconstruction, $\hat{\mathbf{x}}^*_0$.

\paragraph{Ablation probability drop.}
Following \citet{fel2023holistic}, we evaluate how well a concept vector $\mathbf{c}_i$ captures its associated class by ablating the $i$-th entry in $\mathbf{w}_0$ and then classifying the reconstructed sample $\hat{\mathbf{x}}^*_0$ using the trained classifier $h$. The metric is defined as:
\begin{equation} \label{eq:ablation-prob-drop-metric}
[h(\mathbf{x}_0)]_i - [h(\hat{\mathbf{x}}^*_0)]_i.
\end{equation}
We report the average over the entire test-set of ImageNet-500. 

\paragraph{Target probability gain.} Additionally, we analyze the insertion case. For a source concept $\mathbf{c}_i$, and a target concept $\mathbf{c}_j$, we edit $\mathbf{w}_0$ by transferring the weight of the $i$-th entry to the $j$-th entry, leaving the $i$-th entry zeroed. Then the metric is defined as:
\begin{equation}
    [h(\hat{\mathbf{x}}^*_0)]_j - [h(\mathbf{x}_0)]_j
\end{equation}
We randomly select $\approx$ 15\% of the classes in the dataset, and compute the average over all their possible combinations.

\paragraph{Image residual cosine.}
While the aforementioned metrics evaluate how well the coefficient vector captures individual concepts through ablation and insertion, they overlook how these operations affect the remaining concepts.  
In the insertion case, to eliminate the impact of concept \(i\) from \(x_0\) and \(j\) from \(\hat{x}^*_0\), we first compute their residuals with respect to the corresponding concept means:
\[
\mathbf{r}_i(\mathbf{x}_0) = \mathbf{x}_0 - \mathbf{P}_{\bm{\mu}_i}(\mathbf{x}_0), \quad \mathbf{r}_j(\hat{\mathbf{x}}^*_0) = \hat{\mathbf{x}}^*_0 - \mathbf{P}_{\bm{\mu}_j}(\hat{\mathbf{x}}^*_0),
\]
where $ \bm{\mu}_i $ and $ \bm{\mu}_j $ denote the unit-norm mean embeddings of classes $i$ and $j$, respectively, and $\mathbf{P}_{\mathbf{\bm{\mu}}}(\mathbf{x})$ is the projection of $\mathbf{x}$ onto vecotr $\bm{\mu}$. 
The residual cosine similarity is then defined as:
\begin{equation} \label{eq:residual} \frac{\langle \mathbf{r}_i(\mathbf{x}_0)\, , \, \mathbf{r}_j(\hat{\mathbf{x}}^*_0)\rangle}{\lVert \mathbf{r}_i(\mathbf{x}_0)\rVert_2 \, \lVert \mathbf{r}_j(\hat{\mathbf{x}}^*_0)\rVert_2} \end{equation}
\begin{figure*}[t]
    \centering
    \includegraphics[width=0.95\linewidth]{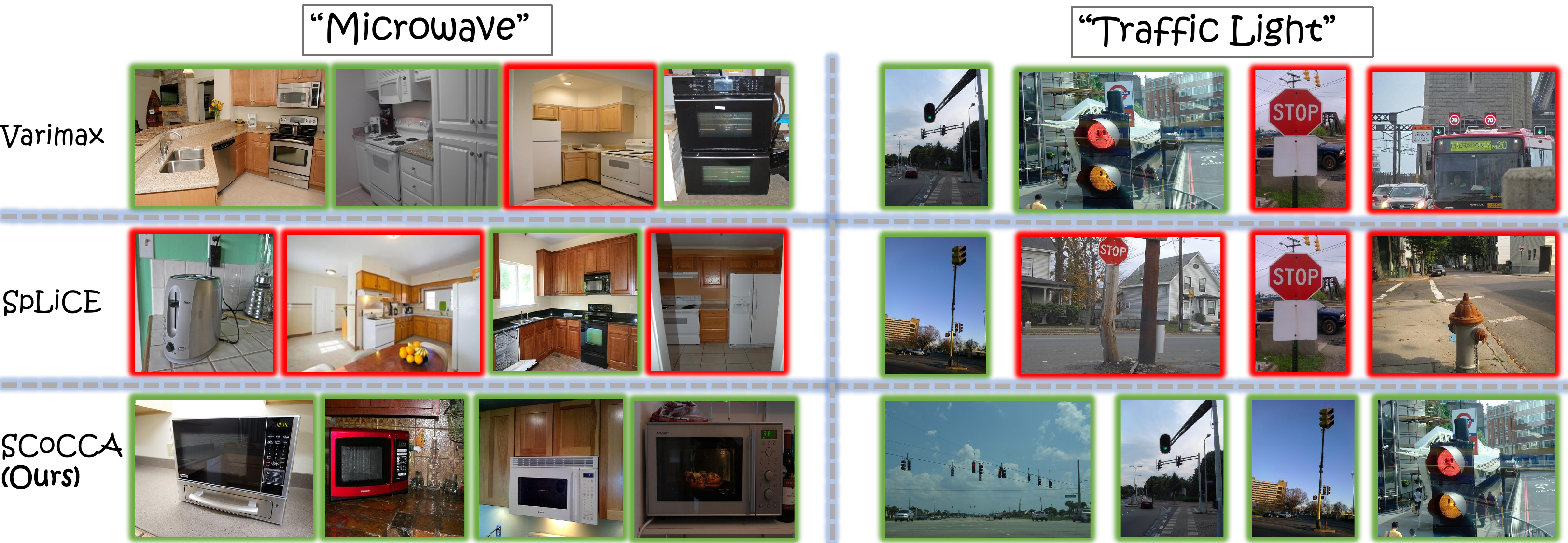}
    \caption{\textbf{Concept Retrieval Generalization.}
    Retrieval of the top four MSCOCO \cite{lin2014microsoft} images with the highest activation for the concepts \emph{Microwave} and \emph{Traffic Light}, where the concept bank was calibrated on ImageNet only (images are shown in descending order). While SCoCCA retrieves images with a clear presence of the concept, other methods in some cases return images with little or no evidence of the desired concept.} 
    \label{fig:microwave}
\end{figure*}

Similar to the target probability gain metric, we randomly select $\approx$ 15\% of the classes in the dataset, and compute the average over all their possible combinations.

\subsubsection{Sparsity}
\paragraph{Concept orthogonality.} The orthogonality metric is defined as:
\begin{equation}
1 - \mathop{\mathbb{E}}_{i \neq j} \left[ \left| \frac{\mathbf{c}_i^\top \mathbf{c}_j}{\lVert \mathbf{c}_i \rVert_2 \,\lVert \mathbf{c}_j \rVert_2} \right| \right]
\end{equation}
It equals $1$ when the concepts are mutually orthogonal and approaches $0$ when concepts are collinear.

\paragraph{Hoyer sparsity.} The Hoyer index \cite{hoyer2004non} is
\begin{equation} \label{eq:hoyer-ch-eval}
 \frac{\sqrt{k} - \lVert \mathbf{w}\rVert_1/\lVert \mathbf{w} \rVert_2}{\sqrt{k} - 1} ,
\end{equation}
where $k$ is the dimension of $\mathbf{w}$. The overall score is the average of the entire test-set.
This scale-invariant index equals $0$ for a constant vector and approaches $1$ when all mass concentrates on a single concept. 
It captures how concentrated the weight distribution is without relying on a hard threshold. 

\paragraph{Energy coverage @10.}  \label{metric:energy-coverage}  
For concept coefficients $\mathbf{w}_i$ we measure how much of the total energy is explained by the 10 most contributing concepts by:
\begin{equation}
\frac{1}{n}\sum_{i=1}^{n}
\frac{\textstyle \sum_{j \in \text{top-}10} (\mathbf{w}_i)_j^2 \lVert \mathbf{\hat{c}}_j\rVert_2^2}
{\textstyle \sum_{j=1}^{k} (\mathbf{w}_i)_j^2 \lVert \mathbf{\hat{c}}_j\rVert_2^2},
\end{equation} 
where $\mathbf{\hat{c}}=\mathbf{c}/\lVert \mathbf{c} \rVert_2$. 
Higher values of this metric indicate that a small number of concepts account for most of the reconstruction energy.

\subsubsection{Reconstruction}
\paragraph{Relative $\ell_2$ reconstruction error.}
This metric quantifies the scale-normalized discrepancy between original and reconstructed embeddings:
\begin{equation}
\label{eq:rel-l2}
\frac{1}{n} \sum_{i=1}^{n}
\frac{\lVert \mathbf{x}_i - \hat{\mathbf{x}}_i \rVert_2^2}
{\lVert \mathbf{x}_i \rVert_2^2}.
\end{equation}
It measures the fraction of signal energy not captured by the reconstruction, providing a scale-invariant complement to the cosine similarity, which focuses on angular alignment.

\paragraph{Cosine reconstruction similarity.}
This metric quantifies the directional consistency between original and reconstructed embeddings:
\begin{equation} \label{eq:cosine-rec}
\frac1n \sum_{i=1}^{n} \frac{\langle \mathbf{x}_i,\hat{\mathbf{x}}_i\rangle}{\lVert \mathbf{x}_i\rVert_2\,\lVert \hat{\mathbf{x}}_i\rVert_2}
\end{equation}
It captures alignment in direction, which is essential for similarity-based retrieval and zero-shot classification methods that rely on normalized embeddings.

\subsection{Results analysis}
The quantitative results across \textit{Purity}, \textit{Editing}, \textit{Sparsity}, and \textit{Reconstruction} metrics are summarized in Tab.~\ref{tab:unified-comparison-bold}. 
SCoCCA consistently outperforms competing approaches, achieving top or comparable performance in nearly all metrics. 
Its advantages are particularly pronounced in purity, editing, and reconstruction, where the improvements over prior dual- and single-modality methods are substantial rather than marginal, highlighting the effectiveness of SCoCCA’s concept decomposition in preserving semantics while enabling precise control. Note that SCoCCA obtains CLIP-level performance on accuracy. 

\paragraph{Purity and concept editing.}
SCoCCA consistently outperforms all baselines across purity-related metrics. It achieves the highest residual cosine similarity ($0.76$), indicating that reconstructed embeddings preserve fine-grained semantic structure without over-smoothing. The method also attains a substantial target probability gain ($0.95$), showing that amplifying a discovered concept reliably strengthens its associated class prediction, and a strong source probability drop ($0.87$). Together, these results demonstrate that SCoCCA achieves high semantic purity while providing precise and disentangled control over concept activations, surpassing both multimodal and single-modality baselines.

\begin{figure}[t]
    \centering
    \includegraphics[width=0.99\linewidth,height=0.28\textheight,keepaspectratio]{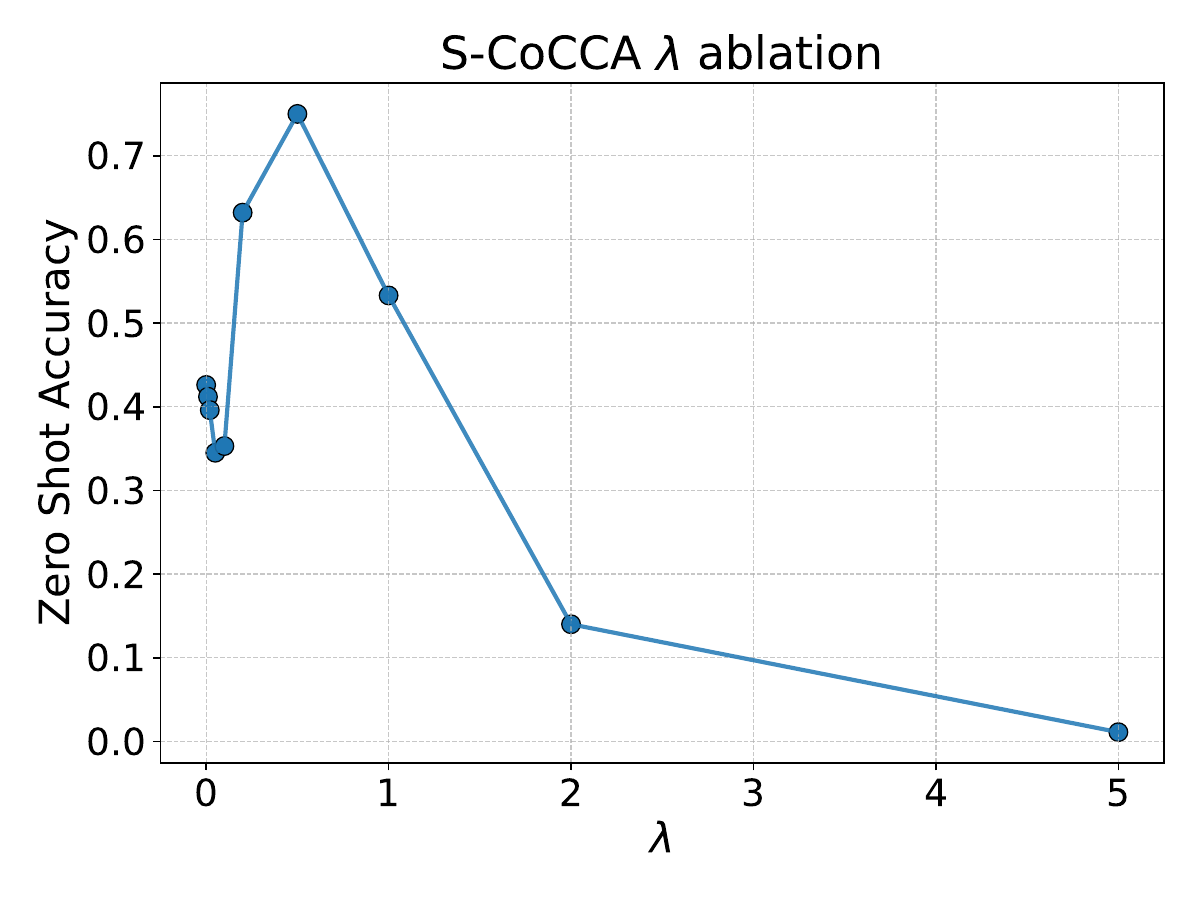} \\[0.2ex]
    \includegraphics[width=0.99\linewidth, height=0.28\textheight,keepaspectratio]{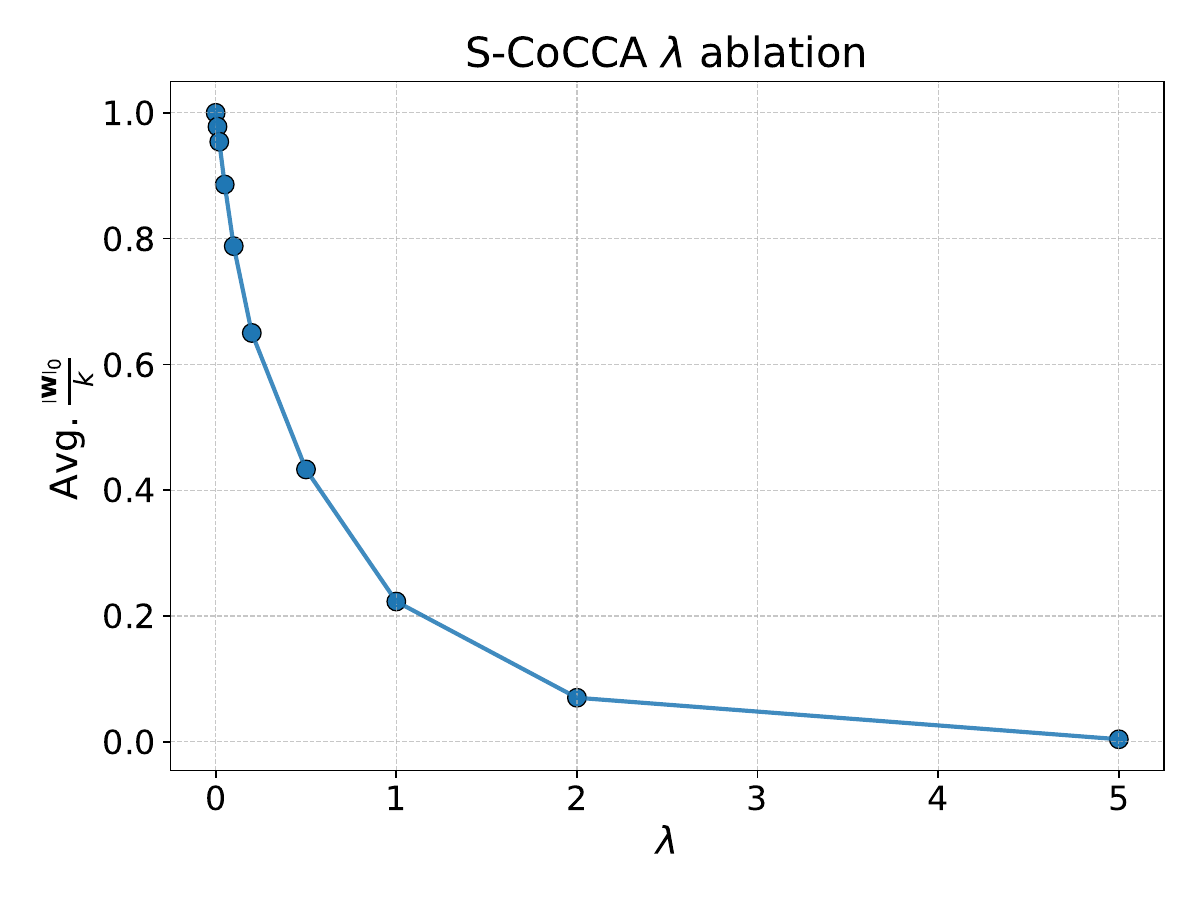}
    \caption{\textbf{$\lambda$ ablation for SCoCCA on ImageNet-500.}\\ \textbf{Top}: Zero-shot accuracy on ImageNet-500 as a function the $\lambda$ used when solving the sparse CoCCA coding objective in \eqref{eq:lasso}. \textbf{Bottom}: $\|w\|_0$ normalized by $k$, as function of $\lambda$. In both plots, $\lambda=0$ corresponds to the CoCCA baseline without the sparsity penalty. The curves shows that forcing sparsity yields substantially higher zero-shot accuracy than the non-sparse baseline.} 
    \label{fig:lambda_ablation}
\end{figure}

\paragraph{Sparsity.}
K-Means inherently produces one-hot concept assignments, yielding perfect $\ell_0$ sparsity by definition. Similarly, Varimax~\cite{zhao2025quantifying} yields orthogonal concept vectors by construction, resulting in ideal scores on the orthogonality metric. While SCoCCA achieves lower orthogonality and sparsity values, incorporating sparsity regularization increases the sparsity of the learned weights, as illustrated in \Cref{fig:lambda_ablation} (bottom).

\paragraph{Reconstruction.}
SCoCCA achieves near-perfect reconstruction fidelity, attaining the highest cosine reconstruction similarity ($0.99$) and the lowest relative $\ell_2$ error ($0.02$), indicating that its decomposed representations can almost perfectly recover the original embeddings while maintaining their directional structure. Furthermore, it attains strong zero-shot accuracy ($0.74$) and precision@5 ($0.85$), on par with or surpassing the CLIP baseline, demonstrating that reconstruction quality directly translates to preserved discriminative performance. These results confirm that SCoCCA’s concept decomposition preserves both geometric integrity and semantic predictiveness of the underlying multimodal space.

\paragraph{Generalization.}
Generalization is a key property of interpretable concept models, as meaningful concepts should extend beyond the dataset on which they were discovered. By leveraging both the image and text representations of CLIP, SCoCCA benefits from the unified multimodal embedding space of a foundation model, which inherently promotes generalization. To evaluate this, we apply SCoCCA’s concept dictionary, calibrated on ImageNet, to the MS-COCO~\cite{lin2014microsoft} validation set and perform a retrieval experiment using a concept query (e.g., \textit{microwave}) to retrieve the top-4 images with the highest concept activations. Two qualitative examples are shown in Fig.~\ref{fig:microwave}, with additional results provided in the Supp. SCoCCA consistently retrieves images that accurately depict the target concept, whereas competing methods often return images where the concept appears only vaguely or is absent. For instance, in the \textit{microwave} query, both SpLiCE and Varimax retrieve images of an entire kitchen, and in the \textit{Traffic Light} query, retrieved images often show general roads or stop signs, sometimes containing only the environment without the concept itself.

\subsection{Ablation Study}
Fig.~\ref{fig:lambda_ablation} demonstrates the zero-shot accuracy as a function of the average fraction of active concepts on ImageNet-500, obtained by varying the sparsity coefficient~$\lambda$, which balances the sparsity and reconstruction terms in Eq.~\ref{eq:lasso}. This measure reflects the relative number of non-zero elements in the weight vector~$\mathbf{w}$, indicating the degree of sparsity. As $\lambda$ increases, sparsity grows accordingly, leading to fewer active concepts. The best performance is achieved when roughly half of the concept weights are zeroed out, suggesting that moderate sparsity yields the most discriminative representations. The ablation plot also includes the case of $\lambda=0$, which corresponds to the CoCCA baseline where no sparsity is enforced. 
\section{Conclusion}
\label{sec:conclusion}
We introduce SCoCCA, a method that bridges cross-modal alignment with concept-based explainability. Built upon Canonical Correlation Analysis (CCA), SCoCCA discovers a shared latent subspace between image and text embeddings while enforcing sparsity for interpretability. Unlike existing concept-based models that rely only on a single modality, SCoCCA aligns multimodal concepts, deriving Concept Activation Vectors (CAVs) that correspond to meaningful semantic directions shared across modalities. By leveraging the CCA objective, SCoCCA implicitly enhances the alignment term of the InfoNCE loss, providing a training-free mechanism to refine pretrained representations such as CLIP. Through extensive experiments, we show that this approach not only enables precise concept decomposition and manipulation but also achieves state-of-the-art performance in reconstruction, as well as in purity and editing metrics, demonstrating that SCoCCA produces both faithful and highly controllable concept representations. Moreover, the learned multimodal concept space generalizes beyond the data used for discovery, supporting retrieval and editing on out-of-distribution images. Taken together, these results indicate that SCoCCA offers a simple and principled way to expose and control the internal conceptual structure of vision-language models, as required for transparent and reliable deployment.
\newpage
{
    \small
    \bibliographystyle{ieeenat_fullname}
    \bibliography{main}

@String(CVPR= {IEEE Conf. Comput. Vis. Pattern Recog.})

@String(AAAI = {AAAI})

@String(CVPR  = {CVPR})

@inproceedings{radford2021learning,
  title={Learning transferable visual models from natural language supervision},
  author={Radford, Alec and Kim, Jong Wook and Hallacy, Chris and Ramesh, Aditya and Goh, Gabriel and Agarwal, Sandhini and Sastry, Girish and Askell, Amanda and Mishkin, Pamela and Clark, Jack and others},
  booktitle={International conference on machine learning},
  pages={8748--8763},
  year={2021},
  organization={PmLR}
}

@article{ghorbani2019towards,
  title={Towards automatic concept-based explanations},
  author={Ghorbani, Amirata and Wexler, James and Zou, James Y and Kim, Been},
  journal={Advances in neural information processing systems},
  volume={32},
  year={2019}
}

@article{fel2023holistic,
  title={A holistic approach to unifying automatic concept extraction and concept importance estimation},
  author={Fel, Thomas and Boutin, Victor and B{\'e}thune, Louis and Cad{\`e}ne, R{\'e}mi and Moayeri, Mazda and And{\'e}ol, L{\'e}o and Chalvidal, Mathieu and Serre, Thomas},
  journal={Advances in Neural Information Processing Systems},
  volume={36},
  pages={54805--54818},
  year={2023}
}

@article{kaiser1958varimax,
  title={The varimax criterion for analytic rotation in factor analysis},
  author={Kaiser, Henry F},
  journal={Psychometrika},
  volume={23},
  number={3},
  pages={187--200},
  year={1958},
  publisher={Springer-Verlag}
}

@inproceedings{lin2014microsoft,
  title={Microsoft coco: Common objects in context},
  author={Lin, Tsung-Yi and Maire, Michael and Belongie, Serge and Hays, James and Perona, Pietro and Ramanan, Deva and Doll{\'a}r, Piotr and Zitnick, C Lawrence},
  booktitle={European conference on computer vision},
  pages={740--755},
  year={2014},
  organization={Springer}
}

@article{hoyer2004non,
  title={Non-negative matrix factorization with sparseness constraints},
  author={Hoyer, Patrik O},
  journal={Journal of machine learning research},
  volume={5},
  number={Nov},
  pages={1457--1469},
  year={2004}
}

@article{mirsky1975trace,
  title={A trace inequality of John von Neumann},
  author={Mirsky, Leon},
  journal={Monatshefte f{\"u}r mathematik},
  volume={79},
  number={4},
  pages={303--306},
  year={1975},
  publisher={Springer}
}

@article{lee2000algorithms,
  title={Algorithms for non-negative matrix factorization},
  author={Lee, Daniel and Seung, H Sebastian},
  journal={Advances in neural information processing systems},
  volume={13},
  year={2000}
}

@article{lloyd1982least,
  title={Least squares quantization in PCM},
  author={Lloyd, Stuart},
  journal={IEEE transactions on information theory},
  volume={28},
  number={2},
  pages={129--137},
  year={1982},
  publisher={IEEE}
}

@inproceedings{andrew2013deep,
  title={Deep canonical correlation analysis},
  author={Andrew, Galen and Arora, Raman and Bilmes, Jeff and Livescu, Karen},
  booktitle={International conference on machine learning},
  pages={1247--1255},
  year={2013},
  organization={PMLR}
}

@article{witten2009penalized,
  title={A penalized matrix decomposition, with applications to sparse principal components and canonical correlation analysis},
  author={Witten, Daniela M and Tibshirani, Robert and Hastie, Trevor},
  journal={Biostatistics},
  volume={10},
  number={3},
  pages={515--534},
  year={2009},
  publisher={Oxford University Press}
}

@article{tibshirani1996lasso,
  author  = {Tibshirani, Robert},
  title   = {Regression Shrinkage and Selection via the Lasso},
  journal = {Journal of the Royal Statistical Society: Series B},
  year    = {1996},
  volume  = {58},
  number  = {1},
  pages   = {267--288}
}

@article{beck2009fista,
  author  = {Beck, Amir and Teboulle, Marc},
  title   = {A Fast Iterative Shrinkage-Thresholding Algorithm for Linear Inverse Problems},
  journal = {SIAM Journal on Imaging Sciences},
  year    = {2009},
  volume  = {2},
  number  = {1},
  pages   = {183--202}
}

@article{Kuhn1955,
  title={The {H}ungarian Method for the Assignment Problem},
  author={Kuhn, Harold W.},
  journal={Naval Research Logistics Quarterly},
  volume={2},
  number={1-2},
  pages={83--97},
  year={1955}
}

@article{parikh2014proximal,
  author  = {Parikh, Neal and Boyd, Stephen},
  title   = {Proximal Algorithms},
  journal = {Foundations and Trends in Optimization},
  year    = {2014},
  volume  = {1},
  number  = {3},
  pages   = {127--239}
}

@inproceedings{levi2025double,
  title     = {The Double Ellipsoid Geometry of CLIP},
  author    = {Levi, Meir Yossef and Gilboa, Guy},
  booktitle = {Proceedings of the 42nd International Conference on Machine Learning},
  series    = {Proceedings of Machine Learning Research},
  volume    = {267},
  year      = {2025},
  publisher = {PMLR},
  address   = {Vancouver, Canada},
}

@inproceedings{betser2025whitenedclip,
  title     = {Whitened CLIP as a Likelihood Surrogate of Images and Captions},
  author    = {Betser, Roy and Levi, Meir Yossef and Gilboa, Guy},
  booktitle = {Proceedings of the 42nd International Conference on Machine Learning},
  series    = {Proceedings of Machine Learning Research},
  volume    = {267},
  year      = {2025},
  publisher = {PMLR},
  address   = {Vancouver, Canada}
}

@article{mairal2014sparse,
  title={Sparse modeling for image and vision processing},
  author={Mairal, Julien and Bach, Francis and Ponce, Jean and others},
  journal={Foundations and Trends{\textregistered} in Computer Graphics and Vision},
  volume={8},
  number={2-3},
  pages={85--283},
  year={2014},
  publisher={Now Publishers, Inc.}
}

@inproceedings{deng2009imagenet,
  title={Imagenet: A large-scale hierarchical image database},
  author={Deng, Jia and Dong, Wei and Socher, Richard and Li, Li-Jia and Li, Kai and Fei-Fei, Li},
  booktitle={2009 IEEE conference on computer vision and pattern recognition},
  pages={248--255},
  year={2009},
  organization={Ieee}
}

@article{kessy2018optimal,
  title={Optimal whitening and decorrelation},
  author={Kessy, Agnan and Lewin, Alex and Strimmer, Korbinian},
  journal={The American Statistician},
  volume={72},
  number={4},
  pages={309--314},
  year={2018},
  publisher={Taylor \& Francis}
}

@article{jendoubi2019whitening,
  title={A whitening approach to probabilistic canonical correlation analysis for omics data integration},
  author={Jendoubi, Takoua and Strimmer, Korbinian},
  journal={BMC bioinformatics},
  volume={20},
  number={1},
  pages={15},
  year={2019},
  publisher={Springer}
}

@misc{SpLiCE2025,
  author = {Bhalla, Usha and Oesterling, Alex and Srinivas, S. and Calmon, Fernando P. and Lakkaraju, Himabindu},
  title = {SpLiCE: Sparse Linear Concept Embeddings},
  howpublished = {GitHub repository},
  year = {2025},
  note = {Available at \url{https://github.com/AI4LIFE-GROUP/SpLiCE/tree/main}}
}

@article{schuhmann2021laion,
  title={Laion-400m: Open dataset of clip-filtered 400 million image-text pairs},
  author={Schuhmann, Christoph and Vencu, Richard and Beaumont, Romain and Kaczmarczyk, Robert and Mullis, Clayton and Katta, Aarush and Coombes, Theo and Jitsev, Jenia and Komatsuzaki, Aran},
  journal={arXiv preprint arXiv:2111.02114},
  year={2021}
}

@inproceedings{NEURIPS2022_702f4db7,
 author = {Liang, Victor Weixin and Zhang, Yuhui and Kwon, Yongchan and Yeung, Serena and Zou, James Y},
 booktitle = {Advances in Neural Information Processing Systems},
 editor = {S. Koyejo and S. Mohamed and A. Agarwal and D. Belgrave and K. Cho and A. Oh},
 pages = {17612--17625},
 publisher = {Curran Associates, Inc.},
 title = {Mind the Gap: Understanding the Modality Gap in Multi-modal Contrastive Representation Learning},
 url = {https://proceedings.neurips.cc/paper_files/paper/2022/file/702f4db7543a7432431df588d57bc7c9-Paper-Conference.pdf},
 volume = {35},
 year = {2022}
}

@InProceedings{pmlr-v119-koh20a,
  title = 	 {Concept Bottleneck Models},
  author =       {Koh, Pang Wei and Nguyen, Thao and Tang, Yew Siang and Mussmann, Stephen and Pierson, Emma and Kim, Been and Liang, Percy},
  booktitle = 	 {Proceedings of the 37th International Conference on Machine Learning},
  pages = 	 {5338--5348},
  year = 	 {2020},
  editor = 	 {III, Hal Daumé and Singh, Aarti},
  volume = 	 {119},
  series = 	 {Proceedings of Machine Learning Research},
  month = 	 {13--18 Jul},
  publisher =    {PMLR},
  url = 	 {https://proceedings.mlr.press/v119/koh20a.html}
}

@article{zhao2025quantifying,
  title={Quantifying Structure in CLIP Embeddings: A Statistical Framework for Concept Interpretation},
  author={Zhao, Jitian and Li, Chenghui and Sala, Frederic and Rohe, Karl},
  journal={arXiv preprint arXiv:2506.13831},
  year={2025}
}

@article{yuksekgonul2022post,
  title={Post-hoc concept bottleneck models},
  author={Yuksekgonul, Mert and Wang, Maggie and Zou, James},
  journal={arXiv preprint arXiv:2205.15480},
  year={2022}
}

@article{kim2023probabilistic,
  title={Probabilistic concept bottleneck models},
  author={Kim, Eunji and Jung, Dahuin and Park, Sangha and Kim, Siwon and Yoon, Sungroh},
  journal={arXiv preprint arXiv:2306.01574},
  year={2023}
}

@inproceedings{moayeri2023text2concept,
  title={Text2concept: Concept activation vectors directly from text},
  author={Moayeri, Mazda and Rezaei, Keivan and Sanjabi, Maziar and Feizi, Soheil},
  booktitle={Proceedings of the IEEE/CVF Conference on Computer Vision and Pattern Recognition},
  pages={3744--3749},
  year={2023}
}

@inproceedings{zhang2021invertible,
  title={Invertible concept-based explanations for cnn models with non-negative concept activation vectors},
  author={Zhang, Ruihan and Madumal, Prashan and Miller, Tim and Ehinger, Krista A and Rubinstein, Benjamin IP},
  booktitle={Proceedings of the AAAI Conference on Artificial Intelligence},
  volume={35},
  number={13},
  pages={11682--11690},
  year={2021}
}

@article{pfau2021robust,
  title={Robust semantic interpretability: Revisiting concept activation vectors},
  author={Pfau, Jacob and Young, Albert T and Wei, Jerome and Wei, Maria L and Keiser, Michael J},
  journal={arXiv preprint arXiv:2104.02768},
  year={2021}
}

@article{bhalla2024interpreting,
  title={Interpreting clip with sparse linear concept embeddings (splice)},
  author={Bhalla, Usha and Oesterling, Alex and Srinivas, Suraj and Calmon, Flavio and Lakkaraju, Himabindu},
  journal={Advances in Neural Information Processing Systems},
  volume={37},
  pages={84298--84328},
  year={2024}
}

@article{amann2020explainability,
  title={Explainability for artificial intelligence in healthcare: a multidisciplinary perspective},
  author={Amann, Julia and Blasimme, Alessandro and Vayena, Effy and Frey, Dietmar and Madai, Vince I and Precise4Q Consortium},
  journal={BMC medical informatics and decision making},
  volume={20},
  number={1},
  pages={310},
  year={2020},
  publisher={Springer}
}

@article{zablocki2022explainability,
  title={Explainability of deep vision-based autonomous driving systems: Review and challenges},
  author={Zablocki, {\'E}loi and Ben-Younes, H{\'e}di and P{\'e}rez, Patrick and Cord, Matthieu},
  journal={International Journal of Computer Vision},
  volume={130},
  number={10},
  pages={2425--2452},
  year={2022},
  publisher={Springer}
}

@article{yacobi2025learning,
  title={Learning Shared Representations from Unpaired Data},
  author={Yacobi, Amitai and Ben-Ari, Nir and Talmon, Ronen and Shaham, Uri},
  journal={arXiv preprint arXiv:2505.21524},
  year={2025}
}

@article{fel2025archetypal,
  title={Archetypal sae: Adaptive and stable dictionary learning for concept extraction in large vision models},
  author={Fel, Thomas and Lubana, Ekdeep Singh and Prince, Jacob S and Kowal, Matthew and Boutin, Victor and Papadimitriou, Isabel and Wang, Binxu and Wattenberg, Martin and Ba, Demba and Konkle, Talia},
  journal={arXiv preprint arXiv:2502.12892},
  year={2025}
}

@article{stevens2025sparse,
  title={Sparse autoencoders for scientifically rigorous interpretation of vision models},
  author={Stevens, Samuel and Chao, Wei-Lun and Berger-Wolf, Tanya and Su, Yu},
  journal={arXiv preprint arXiv:2502.06755},
  year={2025}
}

@article{lim2024sparse,
  title={Sparse autoencoders reveal selective remapping of visual concepts during adaptation},
  author={Lim, Hyesu and Choi, Jinho and Choo, Jaegul and Schneider, Steffen},
  journal={arXiv preprint arXiv:2412.05276},
  year={2024}
}

@article{joseph2025steering,
  title={Steering CLIP's vision transformer with sparse autoencoders},
  author={Joseph, Sonia and Suresh, Praneet and Goldfarb, Ethan and Hufe, Lorenz and Gandelsman, Yossi and Graham, Robert and Bzdok, Danilo and Samek, Wojciech and Richards, Blake Aaron},
  journal={arXiv preprint arXiv:2504.08729},
  year={2025}
}

@incollection{hotelling1992relations,
  title={Relations between two sets of variates},
  author={Hotelling, Harold},
  booktitle={Breakthroughs in statistics: methodology and distribution},
  pages={162--190},
  year={1992},
  publisher={Springer}
}

@inproceedings{kim2018interpretability,
  author = {Kim, Been and Wattenberg, Martin and Gilmer, Justin and Cai, Carrie and Wexler, James and Viégas, Fernanda and Sayres, Rory},
  title = {Interpretability Beyond Feature Attribution: Quantitative Testing with Concept Activation Vectors},
  booktitle = {Proceedings of the 35th International Conference on Machine Learning (ICML)},
  year = {2018}
}

@article{hotelling1936cca,
  title={Relations Between Two Sets of Variates},
  author={Hotelling, Harold},
  journal={Biometrika},
  volume={28},
  number={3/4},
  pages={321--377},
  year={1936}
}

@inproceedings{koh2020concept,
  title={Concept Bottleneck Models},
  author={Koh, Pang Wei and Nguyen, Thao and Tang, Yew Siang and Mussmann, Stephen and Pierson, Emma and Kim, Been and Liang, Percy},
  booktitle={Proceedings of the 37th International Conference on Machine Learning (ICML)},
  year={2020}
}

@inproceedings{chauhan2023interactive,
  title={Interactive Concept Bottleneck Models},
  author={Chauhan, Kushal and Tiwari, Rishabh and Freyberg, Jan and Shenoy, Pradeep and Dvijotham, D.},
  booktitle={Proceedings of the AAAI Conference on Artificial Intelligence (AAAI)},
  year={2023}
}

@inproceedings{vandenhirtz2024stochastic,
  title={Stochastic Concept Bottleneck Models},
  author={Vandenhirtz, Moritz and Laguna, Sonia and Marcinkevičs, Ričards and Vogt, Julia E.},
  booktitle={Proceedings of the 38th Conference on Neural Information Processing Systems (NeurIPS)},
  year={2024}
}

@article{steinmann2023learning,
  title={Learning to intervene on concept bottlenecks},
  author={Steinmann, David and Stammer, Wolfgang and Friedrich, Felix and Kersting, Kristian},
  journal={arXiv preprint arXiv:2308.13453},
  year={2023}
}

@article{sawada2022concept,
  title={Concept bottleneck model with additional unsupervised concepts},
  author={Sawada, Yoshihide and Nakamura, Keigo},
  journal={IEEE Access},
  volume={10},
  pages={41758--41765},
  year={2022},
  publisher={IEEE}
}

@inproceedings{shin2023closer,
  title={A closer look at the intervention procedure of concept bottleneck models},
  author={Shin, Sungbin and Jo, Yohan and Ahn, Sungsoo and Lee, Namhoon},
  booktitle={International Conference on Machine Learning},
  pages={31504--31520},
  year={2023},
  organization={PMLR}
}

@article{sun2024concept,
  title={Concept bottleneck large language models},
  author={Sun, Chung-En and Oikarinen, Tuomas and Ustun, Berk and Weng, Tsui-Wei},
  journal={arXiv preprint arXiv:2412.07992},
  year={2024}
}

@inproceedings{fel2023craft,
  title={Craft: Concept recursive activation factorization for explainability},
  author={Fel, Thomas and Picard, Agustin and Bethune, Louis and Boissin, Thibaut and Vigouroux, David and Colin, Julien and Cad{\`e}ne, R{\'e}mi and Serre, Thomas},
  booktitle={Proceedings of the IEEE/CVF Conference on Computer Vision and Pattern Recognition},
  pages={2711--2721},
  year={2023}
}

@article{zang2024pre,
  title={Pre-trained vision-language models learn discoverable visual concepts},
  author={Zang, Yuan and Yun, Tian and Tan, Hao and Bui, Trung and Sun, Chen},
  journal={arXiv preprint arXiv:2404.12652},
  year={2024}
}

@article{lee2023language,
  title={Language-informed visual concept learning},
  author={Lee, Sharon and Zhang, Yunzhi and Wu, Shangzhe and Wu, Jiajun},
  journal={arXiv preprint arXiv:2312.03587},
  year={2023}
}

@article{li2024vision,
  title={Do vision and language models share concepts? a vector space alignment study},
  author={Li, Jiaang and Kementchedjhieva, Yova and Fierro, Constanza and S{\o}gaard, Anders},
  journal={Transactions of the Association for Computational Linguistics},
  volume={12},
  pages={1232--1249},
  year={2024},
  publisher={MIT Press 255 Main Street, 9th Floor, Cambridge, Massachusetts 02142, USA~…}
}

@article{schrodi2024two,
  title={Two effects, one trigger: On the modality gap, object bias, and information imbalance in contrastive vision-language representation learning},
  author={Schrodi, Simon and Hoffmann, David T and Argus, Max and Fischer, Volker and Brox, Thomas},
  journal={arXiv preprint arXiv:2404.07983},
  year={2024}
}

@article{ramesh2022hierarchical,
  title={Hierarchical text-conditional image generation with clip latents},
  author={Ramesh, Aditya and Dhariwal, Prafulla and Nichol, Alex and Chu, Casey and Chen, Mark},
  journal={arXiv preprint arXiv:2204.06125},
  volume={1},
  number={2},
  pages={3},
  year={2022}
}

@article{oord2018representation,
  title={Representation learning with contrastive predictive coding},
  author={Oord, Aaron van den and Li, Yazhe and Vinyals, Oriol},
  journal={arXiv preprint arXiv:1807.03748},
  year={2018}
}

@inproceedings{wang2020understanding,
  title={Understanding contrastive representation learning through alignment and uniformity on the hypersphere},
  author={Wang, Tongzhou and Isola, Phillip},
  booktitle={International conference on machine learning},
  pages={9929--9939},
  year={2020},
  organization={PMLR}
}

@article{sun2019multi,
  title={Multi-modal sentiment analysis using deep canonical correlation analysis},
  author={Sun, Zhongkai and Sarma, Prathusha K and Sethares, William and Bucy, Erik P},
  journal={arXiv preprint arXiv:1907.08696},
  year={2019}
}

@InProceedings{Hadgi_2025_CVPR,
    author    = {Hadgi, Souhail and Moschella, Luca and Santilli, Andrea and Gomez, Diego and Huang, Qixing and Rodol\`a, Emanuele and Melzi, Simone and Ovsjanikov, Maks},
    title     = {Escaping Plato's Cave: Towards the Alignment of 3D and Text Latent Spaces},
    booktitle = {Proceedings of the IEEE/CVF Conference on Computer Vision and Pattern Recognition (CVPR)},
    month     = {June},
    year      = {2025},
    pages     = {19825-19835}
}

@article{pedregosa2011scikit,
  title={Scikit-learn: Machine learning in Python},
  author={Pedregosa, Fabian and Varoquaux, Ga{\"e}l and Gramfort, Alexandre and Michel, Vincent and Thirion, Bertrand and Grisel, Olivier and Blondel, Mathieu and Prettenhofer, Peter and Weiss, Ron and Dubourg, Vincent and others},
  journal={the Journal of machine Learning research},
  volume={12},
  pages={2825--2830},
  year={2011},
  publisher={JMLR. org}
}
}

\clearpage
\setcounter{page}{1}
\maketitlesupplementary
\section{Ablation of k hyperparameter}

\begin{figure}[H]
    \centering \includegraphics[width=0.99\linewidth,height=0.35\textheight,keepaspectratio]{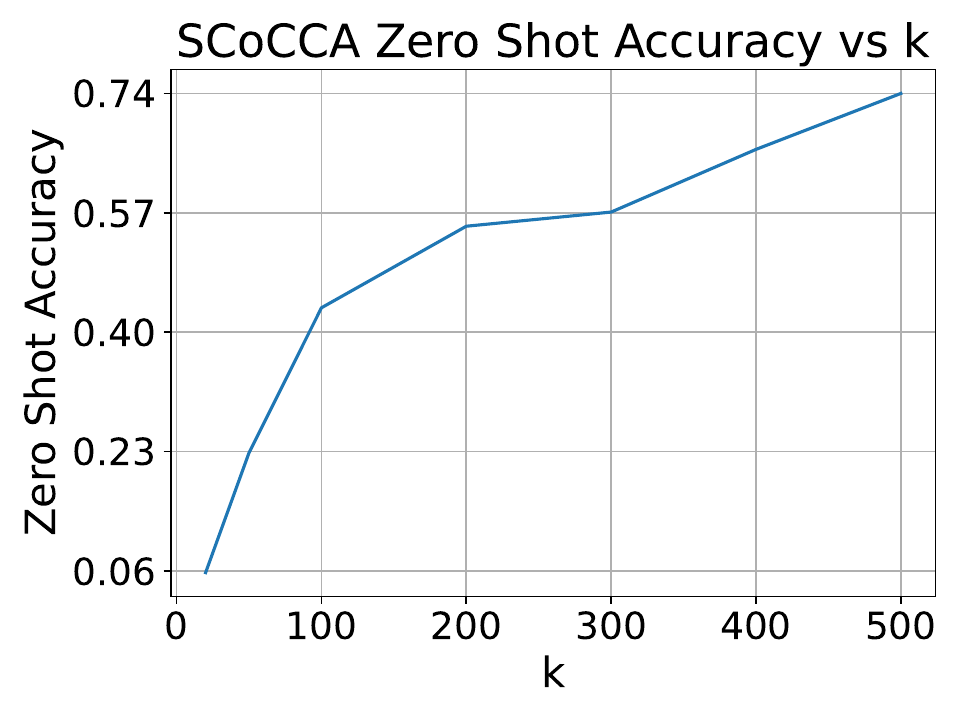} 
    \caption{\textbf{ablation of $k$ hyperparameter} Zero-Shot Accuracy performance on test set of ImageNet-500, as a function of $k$, the number of concepts computed, for the SCoCCA method. As more concepts allows better separation, this is a monotone increasing function. However the rate of steepness decreases, as additional concepts yield less of a return in classification. }
    \label{fig:k_ablation}
\end{figure}

\section{Comparison on CLIP B-32 model}
\begin{table*}[b]
\centering
\caption{\textbf{Comprehensive Performance Comparison.} Results comparing concept decomposition methods on subset of 500 random classes from ImageNet~\cite{deng2009imagenet}, similar to \Cref{tab:unified-comparison-bold}, but \textbf{for CLIP model B-32}. Best results are shown in \textbf{bold}, and second-best are \underline{underlined}. We observe similar trends like \Cref{tab:unified-comparison-bold}. As expected from a smaller model, for same number of $k=500$, reconstruction metrics improve, whereas Purity metrics decrease.}
\footnotesize
\begin{tabular}{l l | cc | c c c c | c}
\toprule
\multirow{2}{*}{Category} & \multirow{2}{*}{Metric} 
& \multicolumn{2}{c|}{Dual-Modality} 
& \multicolumn{4}{c|}{Single-Modality} 
& \multicolumn{1}{c}{Baseline} \\
\cmidrule(lr){3-4} \cmidrule(lr){5-8} \cmidrule(lr){9-9}
 &  & SCoCCA \textbf{(Ours)} & SpLiCE \cite{bhalla2024interpreting} 
 & TCAV\cite{kim2018interpretability} & Varimax\cite{zhao2025quantifying} 
 & NMF & K-Means & CLIP\cite{radford2021learning} \\
 
\midrule
\multirow{5}{*}{\textbf{Purity and Editing}} 
& Ablation prob. drop (\(\uparrow\)) & \underline{0.77} & 0.12 & \textbf{0.79} & 0.73 & 0.00 & 0.37 & - \\
& Target prob gain (\(\uparrow\)) & \textbf{0.72} & 0.11 & 0.02 & \underline{0.61} & -0.19 & -0.39 & - \\
& Img residual cosine (\(\uparrow\)) & \textbf{0.70} & \underline{0.63} & 0.58 & 0.57 & 0.49 & 0.68 & - \\
& Zero-shot accuracy (\(\uparrow\)) & \textbf{0.60} & 0.37 & 0.48 & \underline{0.52} & 0.10 & 0.40 & \textsl{0.63} \\
 & Zero-shot precision@5 (\(\uparrow\)) & \textbf{0.81} & 0.58 & 0.71 & \underline{0.74} & 0.25 & 0.68 & \textsl{0.82} \\
\midrule
\multirow{3}{*}{\textbf{Sparsity}} 
 & Concepts orthogonality (\(\uparrow\)) & \underline{0.92} & \underline{0.92} & 0.81 & \textbf{1.0} & 0.23 & 0.83 & - \\
 & Energy coverage@10 (\(\uparrow\)) & 0.38 & \underline{0.79} & 0.44 & 0.51 & 0.07 & \textbf{1.0} & - \\
 & Hoyer sparsity (\(\uparrow\)) & 0.58 & \underline{0.90} & 0.39 & 0.41 & 0.14 & \textbf{1.0} & - \\
\midrule
\multirow{2}{*}{\textbf{Reconstruction}} 
 & Cosine rec. similarity (\(\uparrow\)) & \textbf{1.00} & 0.69 & 0.64 & \textbf{1.00} & 0.84 & \underline{0.92} & - \\
 & Relative $L_2$ rec. error (\(\downarrow\)) & \textbf{0.00} & 0.22 & 0.11 & \underline{0.02} & 0.40 & 0.27 & - \\
\midrule
\end{tabular}
\label{tab:unified-comparison-bold-b32}
\end{table*}

\clearpage

\section{Derivation of CCA via whitening and SVD}
\subsection{Setup}
Let $ \mathbf{X},\mathbf{Y} \in \mathbb{R}^{n \times d}$
denote paired samples, with rows centered to have zero empirical mean.
Define empirical covariance and cross-covariance matrices:
\begin{equation}
  \bm{\Sigma}_{X} := \frac{1}{n}\,\mathbf{X}^{\top}\mathbf{X} \in \mathbb{R}^{d\times d},\
  \bm{\Sigma}_{Y} := \frac{1}{n}\,\mathbf{Y}^{\top}\mathbf{Y} \in \mathbb{R}^{d\times d},
\end{equation}
and
\begin{equation}
   \bm{\Sigma}_{XY} := \frac{1}{n}\,\mathbf{X}^{\top}\mathbf{Y} \in \mathbb{R}^{d\times d}. 
\end{equation}
Assume $\bm{\Sigma}_{X}$ and $\bm{\Sigma}_{Y}$ are symmetric positive definite, so that their symmetric inverse square roots $\bm{\Sigma}_{X}^{-1/2}$ and $\bm{\Sigma}_{Y}^{-1/2}$ are well defined.

Canonical Correlation Analysis (CCA) with target dimension $k$ solves
\begin{equation}
  \label{eq:cca-obj-matrix2}
  \max_{\mathbf{U},\mathbf{V} \in \mathbb{R}^{d \times k}}
  \ \operatorname{tr}\big[\mathbf{U}^{\top}\bm{\Sigma}_{XY}\mathbf{V}\big]
\end{equation}
subject to the normalization constraints
\begin{equation}
  \label{eq:cca-orthogonal-matrices2}
  \mathbf{U}^{\top}\bm{\Sigma}_{X}\mathbf{U} = \mathbf{I}_{k},
  \qquad
  \mathbf{V}^{\top}\bm{\Sigma}_{Y}\mathbf{V} = \mathbf{I}_{k}.
\end{equation}

\subsection{Whitening and reduction to an orthogonal problem}

Introduce the change of variables
\begin{equation}
  \label{eq:A-B-def}
  \mathbf{A} := \bm{\Sigma}_{X}^{1/2}\mathbf{U},
  \qquad
  \mathbf{B} := \bm{\Sigma}_{Y}^{1/2}\mathbf{V},
\end{equation}
so that
\begin{equation}
  \label{eq:U-V-from-A-B}
  \mathbf{U} = \bm{\Sigma}_{X}^{-1/2}\mathbf{A},
  \qquad
  \mathbf{V} = \bm{\Sigma}_{Y}^{-1/2}\mathbf{B}.
\end{equation}

The constraints \eqref{eq:cca-orthogonal-matrices2} become
\[
\mathbf{A}^{\top}\mathbf{A}
= \mathbf{U}^{\top}\bm{\Sigma}_{X}\mathbf{U}
= \mathbf{I}_{k},
\qquad
\mathbf{B}^{\top}\mathbf{B}
= \mathbf{V}^{\top}\bm{\Sigma}_{Y}\mathbf{V}
= \mathbf{I}_{k}.
\]
Thus $\mathbf{A}$ and $\mathbf{B}$ have orthonormal columns.

Define the whitened cross-covariance matrix
\begin{equation}
  \label{eq:M-def}
  \mathbf{M} := \bm{\Sigma}_{X}^{-1/2}\bm{\Sigma}_{XY}\bm{\Sigma}_{Y}^{-1/2} \in \mathbb{R}^{d\times d}.
\end{equation}
Using \eqref{eq:U-V-from-A-B}, the CCA objective \eqref{eq:cca-obj-matrix2} can be written as
\begin{align}
\operatorname{tr}\big[\mathbf{U}^{\top}\bm{\Sigma}_{XY}\mathbf{V}\big]
  &= \operatorname{tr}\big[\mathbf{A}^{\top}\bm{\Sigma}_{X}^{-1/2}\bm{\Sigma}_{XY}\bm{\Sigma}_{Y}^{-1/2}\mathbf{B}\big] \\
   &= \operatorname{tr}\big[\mathbf{A}^{\top}\mathbf{M}\mathbf{B}\big].
\end{align}

Hence CCA is equivalent to the orthogonal trace maximization
\begin{equation}
  \label{eq:orthogonal-problem}
  \max_{\mathbf{A},\mathbf{B} \in \mathbb{R}^{d \times k}}
  \ \operatorname{tr}\big[\mathbf{A}^{\top}\mathbf{M}\mathbf{B}\big]
\end{equation}
subject to
\begin{equation}
\mathbf{A}^{\top}\mathbf{A} = \mathbf{I}_{k},\  \mathbf{B}^{\top}\mathbf{B} = \mathbf{I}_{k}.    
\end{equation}

\subsection*{SVD of $\mathbf{M}$ and closed form for $\mathbf{U},\mathbf{V}$}

Compute the singular value decomposition of $\mathbf{M}$:
\begin{equation}
  \label{eq:M-SVD}
  \mathbf{M}
  = \mathbf{Q}_{X}\,\mathbf{S}\,\mathbf{Q}_{Y}^{\top},
\end{equation}
where
\begin{itemize}
  \item $\mathbf{Q}_{X},\mathbf{Q}_{Y} \in \mathbb{R}^{d\times d}$ are orthogonal,
        $\mathbf{Q}_{X}^{\top}\mathbf{Q}_{X} = \mathbf{Q}_{Y}^{\top}\mathbf{Q}_{Y} = \mathbf{I}_{d}$,
  \item $\mathbf{S} = \operatorname{diag}(s_{1},\dots,s_{d})$ with singular values
  \begin{equation}
      s_{1} \geq s_{2} \geq \dots \geq s_{d} \geq 0.
  \end{equation}
\end{itemize}

Let $\mathbf{Q}_{X,k} \in \mathbb{R}^{d\times k}$ and $\mathbf{Q}_{Y,k} \in \mathbb{R}^{d\times k}$ denote the matrices formed by the first $k$ columns of $\mathbf{Q}_{X}$ and $\mathbf{Q}_{Y}$ respectively, and let
\[
\mathbf{S}_{k} := \operatorname{diag}(s_{1},\dots,s_{k}) \in \mathbb{R}^{k\times k}.
\]

A standard result (von Neumann's trace inequality \cite{mirsky1975trace}) implies that the solution of \eqref{eq:orthogonal-problem} is obtained by taking
\begin{equation}
  \label{eq:Astar-Bstar}
  \mathbf{A}^{\star} = \mathbf{Q}_{X,k},\qquad
  \mathbf{B}^{\star} = \mathbf{Q}_{Y,k},
\end{equation}
up to a common right multiplication by an orthogonal matrix in $\mathbb{R}^{k\times k}$.
For this choice,
\begin{equation}
  \label{eq:orthogonal-opt-value}
  \operatorname{tr}\big[(\mathbf{A}^{\star})^{\top}\mathbf{M}\mathbf{B}^{\star}\big]
  = \operatorname{tr}(\mathbf{S}_{k})
  = \sum_{i=1}^{k} s_{i},
\end{equation}
which is the maximum possible value of \eqref{eq:orthogonal-problem}.

Back-substituting \eqref{eq:Astar-Bstar} into \eqref{eq:U-V-from-A-B}, we obtain the CCA projection matrices in closed form:
\begin{equation}
  \label{eq:U-V-closed-form}
  \mathbf{U}^{\star} = \bm{\Sigma}_{X}^{-1/2}\mathbf{Q}_{X,k},
  \qquad
  \mathbf{V}^{\star} = \bm{\Sigma}_{Y}^{-1/2}\mathbf{Q}_{Y,k}.
\end{equation}

\section{Methods} \label{subsec:methods}
\paragraph{Varimax} In a recent application by \citet{zhao2025quantifying}, Varimax seeks an orthogonal rotation that concentrates loadings for interpretability. The method uses a $k$-truncated PCA to decompose centered image embeddings $\mathbf{X}$ into $(\mathbf{U}_k  \mathbf{D}_k)\mathbf{V}_k^{\top}$. It then computes an orthogonal rotation $\mathbf{R}$ that maximizes the Varimax objective \cite{kaiser1958varimax}.
From these the authors compute a concept bank $\mathbf{C} = \mathbf{V}_k \mathbf{R}$ and coefficients $\mathbf{W} = (\mathbf{U}_k \mathbf{D}_k) \mathbf{R}$. 

\paragraph{SpLiCE} \cite{bhalla2024interpreting} sets a concept dictionary $\mathbf{C}$ as the CLIP text embeddings of a vocabulary constructed from the 15000 most frequent one- and two-word bigrams in the text captions of the LAION-400m dataset \cite{schuhmann2021laion}. In addition, for each dataset, the names of the classes of that dataset are added to the vocabulary, as described in the paper. Using the concept dictionary $\mathbf{C}$, SpLiCE decomposes an image $x$ by solving the LASSO equation given by \eqref{eq:lasso}. By varying $\lambda$, different sparsity-reconstruction tradeoffs can be achieved. 

\paragraph{K-Means} For this (baseline) method, We compute the $k$-means \cite{lloyd1982least} over the centered image embeddings $\mathbf{X}$. The vectors to the computed centroids are used as the concept vectors. For a new image embedding $\mathbf{x}_0$, the coefficient $\mathbf{w}_0$ is the one-hot vector pointing to the nearest centroid. 

\paragraph{NMF} We use Nonnegative Matrix Factorization. As common, we shift $\mathbf{X}$ to be non-negative, by computing a scalar shift $s = \min(0, \displaystyle \min_{i,j} (\mathbf{X}_{ij}))$, so $s \leq 0$, and then shifting $\mathbf{X}_{sh} = \mathbf{X} - s\mathbf{1}\mathbf{1}^{\top}$ so $\mathbf{X}_{sh} \geq 0$ entry-wise.
We then solve the optimization problem 
\begin{equation}
\displaystyle \min_{\mathbf{W} \geq 0, \mathbf{C} \geq 0} \frac{1}{2}   \lVert \mathbf{X}_{sh}^T - \mathbf{C} \mathbf{W} \rVert_F^2
\end{equation}
using the multiplicative-updates (MU) solver of sklearn \cite{lee2000algorithms}. For reconstruction, we un-shift by adding $s$ back.

\paragraph{TCAV} \cite{kim2018interpretability} works in a supervised setting. Let $f$ be a model providing activations from images. For a concept $c$, the user provides images $P_c$ that represent that concept (e.g. ``striped'') and a negative set $N$ of random images. A binary linear classifier is trained to separate activations of the positive set $\{ f(\mathbf{x}) \mid \mathbf{x} \in P_c \}$ from activations of the negative set $ \{ f(\mathbf{x}) \mid \mathbf{x} \in N \} $. The normal to the hyperplane separating the activations is used as that Concept Activation Vector (CAV) for concept $c$. For $k$ classes we compute $\mathbf{C} \in \mathbb{R}^{d \times k}$. For comparison, coefficients $\mathbf{W}$ are computed by solving the LASSO equation \eqref{eq:lasso}.

\end{document}